\documentclass{article}

\usepackage{PRIMEarxiv}

\usepackage[utf8]{inputenc} % allow utf-8 input
\usepackage[T1]{fontenc}    % use 8-bit T1 fonts
\usepackage{hyperref}       % hyperlinks
\usepackage{url}            % simple URL typesetting
\usepackage{booktabs}       % professional-quality tables
\usepackage{amsfonts}       % blackboard math symbols
\usepackage{nicefrac}       % compact symbols for 1/2, etc.
\usepackage{microtype}      % microtypography
\usepackage{lipsum}
\usepackage{fancyhdr}       % header
\usepackage{graphicx}       % graphics
\graphicspath{{media/}}     % organize your images and other figures under media/ folder
\usepackage{amsmath}
\usepackage{subfigure}

\title{Brain Network Classification Based on Graph Contrastive Learning and Graph Transformer
\thanks{\textit{\underline{}}
\textbf{Equal Contribution}} 
}

\author{
  Lan,YAO \\
  School of Mathematics \\
  University of Hunan \\
  Changsha\\
  \texttt{yao@hnu.edu.cn} \\
   \And
  ZhiTeng,ZHU \\
  School of Mathematics \\
  University of Hunan \\
  Changsha\\
  \texttt{fxl158521@hnu.edu.cn} \\
}

\begin{document}
\maketitle

\begin{abstract}
The dynamic characterization of functional brain networks is of great significance for elucidating the mechanisms of human brain function. Although graph neural networks have achieved remarkable progress in functional network analysis, challenges such as data scarcity and insufficient supervision persist. To address the limitations of limited training data and inadequate supervision, this paper proposes a novel model named PHGCL-DDGformer that integrates graph contrastive learning with graph transformers, effectively enhancing the representation learning capability for brain network classification tasks. To overcome the constraints of existing graph contrastive learning methods in brain network feature extraction, an adaptive graph augmentation strategy combining attribute masking and edge perturbation is implemented for data enhancement. Subsequently, a dual-domain graph transformer (DDGformer) module is constructed to integrate local and global information, where graph convolutional networks aggregate neighborhood features to capture local patterns while attention mechanisms extract global dependencies. Finally, a graph contrastive learning framework is established to maximize the consistency between positive and negative pairs, thereby obtaining high-quality graph representations. Experimental results on real-world datasets demonstrate that the PHGCL-DDGformer model outperforms existing state-of-the-art approaches in brain network classification tasks.
\end{abstract}

% keywords can be removed

\section{Introduction}
The brain, as a highly complex system, has long been a central focus of modern neuroscience in understanding its functional organization. Resting-state functional magnetic resonance imaging (rsfMRI) technology, which measures time-varying changes in blood oxygen level-dependent (BOLD) signals, provides a powerful tool for revealing brain activity. As a product of fMRI data, the functional connectome (FC) captures the synchronization patterns of BOLD signals across different brain regions, serving as a reliable indicator for decoding functional variability in the brain. It plays a crucial role in cognitive neuroscience, medical research, and clinical applications \cite{Grabner}\cite{Bullmore2009}\cite{Finn}\cite{Fernandes}.

In recent years, the rapid development of deep learning techniques, particularly the application of networks such as LSTM \cite{Hong} and CNN \cite{Ji} in functional brain network analysis, has opened new possibilities for extracting deep-level features and interaction relationships from networks \cite{Eslami}. Compared to traditional machine learning methods \cite{Guo}, deep models are better equipped to extract more discriminative features from brain networks. The application of graph convolutional networks (GCN) on graph data has inspired novel approaches for brain network representation learning. The graph-theoretical model of brain networks, which defines anatomical regions as nodes and functional connections as edges, effectively quantifies functional connectomes (FCs) and reveals their neurobiological significance \cite{Wang}. To deepen the understanding of relationships among functional connections, various graph neural network (GNN) variants have been proposed, such as BrainGNN \cite{Li}, MVN-GCN \cite{Wen1}, BrainGSL \cite{Wen2}, FSL-BrainNet \cite{LiLanting}, STAGIN \cite{Kim}, and ST-GCN \cite{Gadgil}. These models enhance the depth and breadth of brain network analysis through different mechanisms. For instance, BrainGNN employs region selection pooling layers and regularization terms to highlight critical brain regions; MVN-GCN leverages multi-view information to improve link prediction accuracy; while STAGIN and ST-GCN incorporate spatiotemporal features to learn dynamic brain networks. These models provide innovative tools and methods for understanding brain function and diagnosing neurological disorders.

Despite significant progress in the application of graph neural networks to functional brain network analysis, several key challenges remain. The core hypothesis for brain disease classification is that individuals with the same type of brain disease exhibit similar structural patterns in graph-based brain network modeling. Based on this assumption, researchers have developed complex supervised graph deep learning models to extract discriminative features for brain network classification. However, the scarcity of brain network data and the resulting insufficient supervision signals have increasingly become a major bottleneck limiting the performance of supervised models in brain network classification. To address this issue, graph self-supervised learning has been proposed to augment data and perform deep-level mining, thereby learning more precise brain network features to enhance model classification performance and generalization capability.

To tackle the challenges faced by graph neural networks in brain network analysis, inspired by the GraphCL model \cite{2024GraphCL}, this chapter proposes a novel graph contrastive learning model—PHGCL-DDGformer (Persistent Homology Graph Contrastive Learning and Dual-domain Graph Transformer Models for Brain Networks)—for brain network classification. GraphCL is a graph contrastive learning model based on data augmentation, which generates graph data with different views through various graph data augmentation methods, thereby constructing a graph contrastive learning framework. Contrastive learning is then employed to learn high-quality graph representations by maximizing consistency between positive and negative pairs. Traditional graph neural networks rely on message-passing mechanisms that effectively encode only local neighborhood information, making it difficult to capture long-range dependencies in graph structures. Conventional graph augmentation methods often focus solely on local node and edge information, lacking adaptive adjustment capabilities for graph features and structures, as well as higher-order structural information such as subgraph patterns, network motifs, and overall topological properties. These limitations hinder the full exploration of a graph's potential value when dealing with complex and diverse graph data. To address these challenges, the proposed PHGCL-DDGformer model aims to significantly enhance the effectiveness and efficiency of graph representation learning while extracting both local and global information from brain network graphs, as well as their complex interaction patterns. In summary, the contributions of this study are as follows:
\begin{itemize}
\item[$\bullet$] Introduction of an attribute masking mechanism that evaluates feature dimension importance based on the absolute values of features, ranks feature dimensions, and masks less important ones. Edge perturbation, on the other hand, measures edge importance through node centrality, thereby deleting edges to disrupt graph structure and enhance the model's understanding and generalization capability for graph structures.

\item[$\bullet$] The DDGformer module extracts local and global information of graph structures by stacking encoders. This module addresses the limitations of GNNs in handling long-range dependencies and oversmoothing, as well as the issue of over-globalization in homogeneous brain networks.

\item[$\bullet$] The contrastive learning module jointly incorporates encoding information and topological information, enabling the model to more comprehensively understand the complex interactions among brain regions.

\item[$\bullet$] The effectiveness of the PHGCL-DDGformer model is successfully validated on the Autism Brain Imaging Data Exchange (ABIDE) and Attention Deficit Hyperactivity Disorder (ADHD) datasets. This study provides an experimentally verified brain network classification model along with detailed results of data processing and analysis.
\end{itemize}

\section{Preliminaries}
A graph is a form of unstructured data composed of nodes and edges, widely applied in various domains including social networks, recommendation systems, protein-protein interaction networks, and knowledge graphs. Graphs effectively represent entities and their complex relationships while revealing the underlying structure of data. In this context, Graph Neural Networks (GNNs), as the core technology of graph deep learning, have become a crucial research direction in this field by processing graph data through deep learning methods. The fundamental principle of GNNs involves iteratively updating node feature representations through information propagation between adjacent nodes, thereby learning both local and global graph information at multiple levels. This approach can generate graph representations suitable for various tasks such as node classification, graph classification, and link prediction. 

Graph Transformers represent an extended variant of GNNs that integrate the self-attention mechanism from Transformer architectures. This design avoids introducing any structural bias in intermediate layers while effectively capturing interaction information between arbitrary node pairs to enhance the expressive power of graph data. Unlike conventional approaches that only capture local neighborhood information, this variant can dynamically acquire global information. The development of Graph Transformers was inspired by the successful applications of Transformers in Natural Language Processing (NLP) and Computer Vision (CV), combined with the structural characteristics of graphs to incorporate graph inductive biases for effective processing of graph data, such as prior knowledge or assumptions about graph attributes. For instance, SGFormer \cite{ren2023sg} employs a simplified global attention mechanism and introduces GNNs to capture local graph information, thereby achieving fusion of local and global information while maintaining efficient generalization to large-scale network datasets. Gradformer \cite{pei2023gradformer} implements an exponentially decaying mask to maintain global awareness while focusing more on local structural information. CoBFormer \cite{xing2024less} addresses the over-globalization problem by utilizing a two-level global attention mechanism to effectively approximate global attention and achieve better classification performance.

The contrastive learning strategy is fundamentally based on the principle of Mutual Information (MI) maximization, where learning occurs by predicting the similarity between two augmented graph instances. For graph-level tasks, same-granularity contrast is typically applied to graph representations and can be formalized as the following optimization problem:
\[
u^*, f^* = \arg\min_{u, f} \mathcal{L}_{\text{con}}(f(\tilde{g}^{(1)}), f(\tilde{g}^{(2)}))
\]
where \(\tilde{g}^{(i)} = R(u(\tilde{A}^{(i)}, \tilde{X}^{(i)}))\) represents the representation of the augmented graph \(\tilde{G}^{(i)}\), and \(R(\cdot)\) is a readout function that generates graph-level embeddings from node representations.

Persistent Homology is a fundamental concept in Topological Data Analysis (TDA) \cite{Rote} with broad applications in time series data analysis. As an emerging research field, TDA has demonstrated significant potential across multiple domains, particularly in time series modeling and feature extraction. In climate analysis, TDA has been used to reveal periodic patterns and trends in climate data. Through persistence analysis of point clouds, it can track the stability of dynamic systems across different time windows, thereby supporting the modeling of complex systems. In periodic analysis of time series, maximum persistence has been employed to quantify periodic features of time series, providing more accurate results than traditional methods \cite{Gakhar}. Furthermore, TDA finds extensive applications in signal processing; for example, the delay coordinate embedding method can effectively detect harmonic structures in respiratory sound signals, which is crucial for automated detection of respiratory abnormalities \cite{Perea}. TDA has also been applied to clustering and classification of time series data, where extracting topological features and feeding them into deep learning models can significantly improve classification performance, particularly demonstrating stronger robustness than traditional methods when handling complex time series data. In motion sensor data analysis, TDA has been used for classifying daily activities and sports activities, capturing subtle differences between different activities to improve classification accuracy \cite{Adams}. In the financial domain, TDA has been employed to analyze and predict market dynamics \cite{Gidea}, offering a novel perspective for identifying market changes and anomalous behaviors through topological feature modeling of financial time series.

Persistent Homology theory is a topological data analysis technique that captures topological features of data and their persistence across different scales using algebraic topological methods, identifying and quantifying structures such as connected components, loops, and voids in the data. Given a dataset \( P \) or topological space \( X \), an increasing filtration of simplicial complexes is constructed: \(\emptyset = X_1 \subseteq X_2 \subseteq \cdots \subseteq X_m = X\), where each \( X_i \) is a simplicial complex formed by progressively adding points or edges. For each complex \( X_i \), we can compute its simplicial homology group \( H_n(X_i) \), representing the "hole" structures present in the space. During this process, homology classes \( [c] \in H_n(X_i) \) appear and disappear as the complex evolves. The "birth time" of a homology class is when it first appears in complex \( X_i \), while its "death time" is when it last appears in complex \( X_j \). The "persistence" of a homology class is defined as the difference between its birth and death times: \(\text{pers}(c) = d_i - b_i\), where \( b_i \) and \( d_i \) are the birth and death times of the homology class, respectively. The persistence information of all homology classes can be encoded in a "persistence diagram": \(D = \{(b_1, d_1), (b_2, d_2), \ldots, (b_r, d_r)\}\), where each point \( (b_i, d_i) \) represents the birth and death times of a homology class.

\section{The PHGCL-DDGformer Model}
\begin{figure}[htbp]
    \centering
    \includegraphics[width=0.8\textwidth]{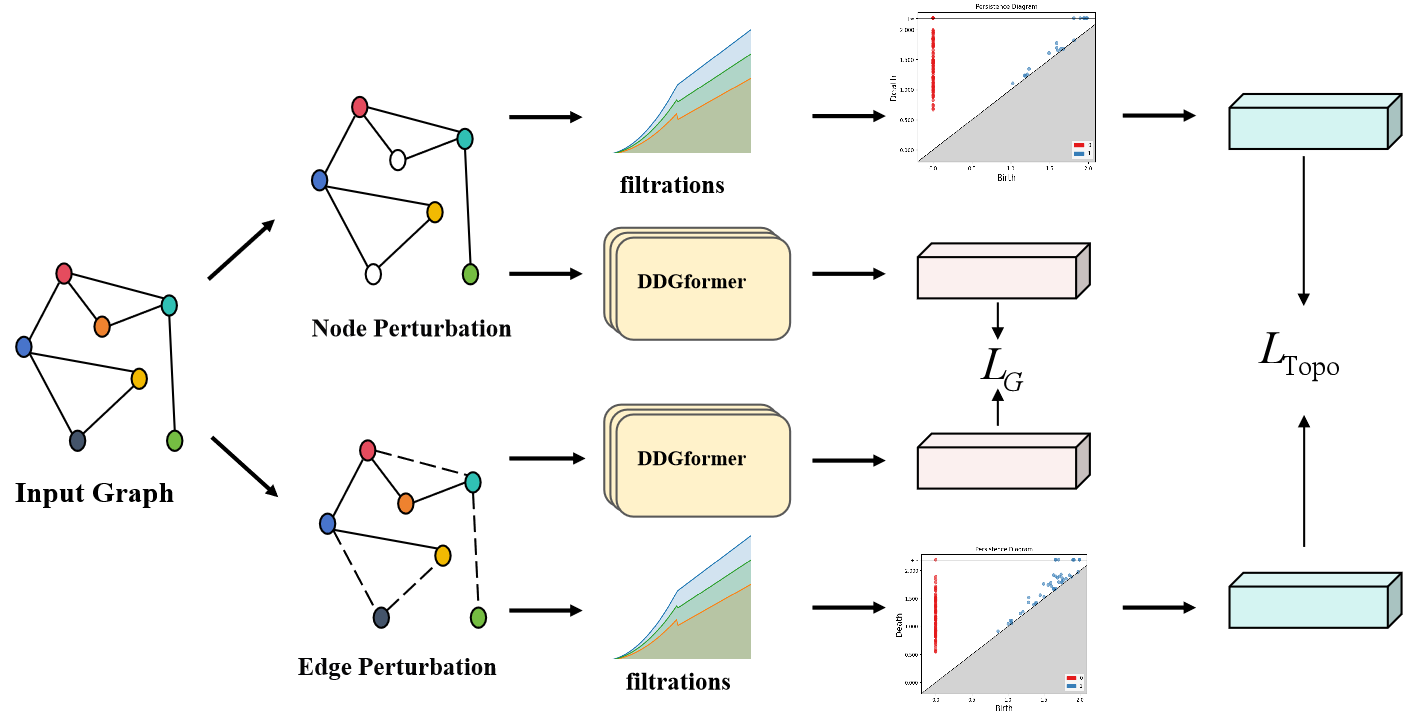}
    \caption{Framework of the PHGCL-DDGformer model}
    \label{fig:a1}
\end{figure}

The framework of the PHGCL-DDGformer model is illustrated in Figure \ref{fig:a1}, which consists of three main modules: (1) Graph Augmentation Module; (2) Dual-Domain Graph Transformer Encoder Module; and (3) Graph Contrastive Learning Module.

\subsection{Graph Augmentation via Importance-Driven Perturbations}

We propose a dual perturbation strategy comprising edge-wise structural perturbation and node-attribute masking, guided by importance scoring mechanisms to preserve essential graph characteristics. The augmentation operates on an attributed graph \(\mathcal{G} = (\mathcal{V}, \mathcal{E}, A, X)\) with node set \(\mathcal{V}\), edge set \(\mathcal{E}\), adjacency matrix \(A\), and feature matrix \(X \in \mathbb{R}^{|\mathcal{V}| \times d_F}\).

\textbf{Edge Perturbation.} We first quantify edge importance through node centrality fusion. For edge \((u,v) \in \mathcal{E}\), define edge centrality as \(\mathbf{w}_{uv}^{e} = \frac{\varphi_c(u) + \varphi_c(v)}{2}\), where \(\varphi_c(\cdot)\) represents the PageRank centrality computed via:
\begin{equation}
    \mathrm{PR}(v) = \frac{1 - d}{|\mathcal{V}|} + d \sum_{u \in \mathcal{N}(v)} \frac{\mathrm{PR}(u)}{|\mathcal{N}(u)|}
    \label{eq:pagerank}
\end{equation}
We then compute normalized removal probabilities through logarithmic scaling \(s_{uv}^e = \log \mathbf{w}_{uv}^e\) and piecewise linear transformation:
\begin{equation}
    p_{uv}^e = \min\left( \frac{s_{\max}^e - s_{uv}^e}{s_{\max}^e - \mu_s^e} \cdot p_e, p_\tau \right)
    \label{eq:edge_prob}
\end{equation}
where \(p_e\) controls the base removal rate, \(s_{\max}^e\) and \(\mu_s^e\) denote maximum and mean scaled centrality, and \(p_\tau\) prevents excessive deletion. The perturbed edge set \(\widetilde{\mathcal{E}}\) is sampled via Bernoulli trials with survival probability \(1 - p_{uv}^e\).

\textbf{Attribute Masking.} For feature dimension \(i\), we calculate importance weights by integrating feature magnitude and node centrality:
\begin{equation}
    \mathbf{w}_i^f = \sum_{u \in \mathcal{V}} |x_{ui}| \cdot \varphi_c(u)
    \label{eq:feat_weight}
\end{equation}
After logarithmic scaling \(s_i^f = \log \mathbf{w}_i^f\), mask probabilities follow:
\begin{equation}
    p_i^f = \min\left( \frac{s_{\max}^f - s_i^f}{s_{\max}^f - \mu_s^f} \cdot p_f, p_\tau \right)
    \label{eq:feat_prob}
\end{equation}
where \(p_f\) regulates the baseline masking rate. The masked feature matrix \(\hat{X}\) is generated via:
\begin{equation}
    \hat{X} = \begin{bmatrix}
        \boldsymbol{x}_1 \circ \boldsymbol{m} \\ 
        \vdots \\ 
        \boldsymbol{x}_N \circ \boldsymbol{m}
    \end{bmatrix}, \quad \boldsymbol{m} \sim \prod_{i=1}^{d_F} \mathrm{Bern}(1-p_i^f)
    \label{eq:masked_feat}
\end{equation}

This dual strategy generates two augmented views \((\mathcal{V}, \widetilde{\mathcal{E}}, A, X)\) and \((\mathcal{V}, \mathcal{E}, A, \hat{X})\) with structure-feature co-perturbation, where hyperparameters \(p_e\) and \(p_f\) control perturbation intensities while centrality-aware scoring preserves semantically crucial patterns. The logarithmic scaling and probability truncation prevent over-destruction of graph topology and feature semantics.

\subsection{Dual-Domain Graph Transformer Encoder}
\begin{figure}[htbp]
    \centering
    \includegraphics[width=0.7\textwidth]{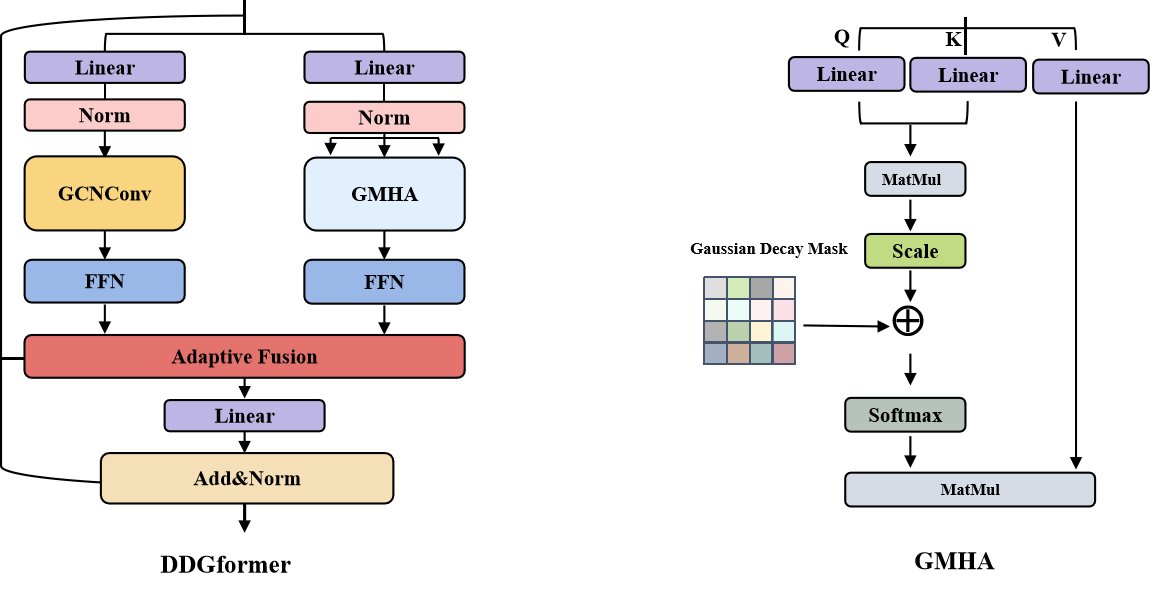}
    \caption{Architecture of the proposed DDGformer encoder}
    \label{fig:encoder_arch}
\end{figure}

The DDGformer module addresses the key challenges in graph representation learning by simultaneously resolving three fundamental limitations: (1) GNNs' inability to capture long-range dependencies, (2) over-smoothing in deep architectures, and (3) over-globalization in homogeneous brain networks. As illustrated in Figure~\ref{fig:encoder_arch}, our architecture integrates graph convolutional operations with a novel Gaussian-masked graph transformer through an adaptive fusion mechanism.

Given an input graph $\mathcal{G}=(\mathcal{V}, \mathcal{E}, A, X)$, we initialize node features as $H^{(0)} = \text{Norm}(\text{Linear}(X))$. The local feature extraction employs graph convolution:

\begin{equation}
H_{\text{gcn}}^{(l+1)} = \text{FFN}(\sigma(\tilde{D}^{-\frac{1}{2}} \tilde{A} \tilde{D}^{-\frac{1}{2}} H^{(l)} W_1))
\end{equation}

where FFN denotes a feed-forward network, $\sigma$ is the activation function, and $W_1$ contains learnable parameters. For global attention, we introduce Gaussian decay masking (GMHA) to preserve structural relationships:

\begin{equation}
 S^{(l+1)} = \frac{H^{(l)}W_2(H^{(l)}W_3)^\mathrm{T}}{\sqrt{d}}, \quad M_{ij} = \exp\left(-\frac{(\psi(v_i, v_j) - \mu)^2}{2\sigma^2}\right), \quad \tilde{S} = \text{softmax}(S \odot M) 
\end{equation}

Here, $\psi(v_i, v_j)$ computes the shortest path distance between nodes, with $\mu$ and $\sigma$ representing its mean and standard deviation respectively. The attention outputs are processed as: \( H_{\text{gt}}^{(l+1)} = \text{FFN}(\tilde{S}H^{(l)}W_4) \), and our adaptive fusion mechanism dynamically balances these components: \( \alpha^{(l+1)} = \sigma(\text{linear}(\text{Mean}(H^{(l)}))) \).yielding the final node representation:

\begin{equation}
H^{(l+1)} = \alpha^{(l+1)} \odot H_{\text{gcn}}^{(l+1)} + (1-\alpha^{(l+1)}) \odot H_{\text{gt}}^{(l+1)}
\end{equation}

For graph-level tasks, node features are aggregated through a readout function:\(h_{\mathcal{G}} = \text{Readout}(H^{(N)})\).The unified architecture processes both node-masked ($\mathcal{G}_F$) and edge-masked ($\mathcal{G}_E$) graphs through shared DDGformer blocks, generating comprehensive representations that capture hierarchical structural information. This design enables simultaneous learning of local topological patterns and global graph properties while maintaining spatial relationships through the Gaussian attention mask.

\subsection{Graph Contrastive Learning}

Our framework employs multi-view contrastive learning across feature and topological spaces. For an input batch of graphs $\mathcal{G} = \{\mathcal{G}_1, \ldots, \mathcal{G}_T\}$, we generate two augmented views per graph $\mathcal{G}_i$ through edge perturbation $\mathcal{G}_{i_E}$ and node attribute masking $\mathcal{G}_{i_F}$. The shared DDGformer encoder processes both views to obtain latent representations $H_{i_E}^+ = \text{DDGformer}(\mathcal{G}_{i_E})$ and $H_{i_F}^+ = \text{DDGformer}(\mathcal{G}_{i_F})$. We optimize feature-level invariance using InfoNCE loss:

\begin{equation}
\mathcal{L}_G = -\frac{1}{T}\sum_{i=1}^T \log \frac{\exp(s(H_{i_E}^+, H_{i_F}^+)/\tau)}{\sum_{k\neq i} \exp(s(H_{i_F}^+, H_{k_F}^-)/\tau)}
\label{eqa:1}
\end{equation}

where $s(u,v) = u^\top v/\|u\|\|v\|$ computes cosine similarity and $\tau$ controls the temperature. For topological consistency, we first compute edge centrality via Eq.~\ref{eqa:1}, then extract persistence diagrams through lower-star filtration, yielding vectorized topological descriptors $To_{i_E}^+$ and $To_{i_F}^+$. The topological contrast loss follows:

\begin{equation}
\mathcal{L}_{Topo} = -\frac{1}{T}\sum_{i=1}^T \log \frac{\exp(s(To_{i_E}^+, To_{i_F}^+)/\tau)}{\sum_{k\neq i} \exp(s(To_{i_F}^+, To_{k_F}^-)/\tau)}
\end{equation}

The complete objective combines supervised classification with contrastive losses:

\begin{equation}
\label{eqqqq}
\mathcal{L} = \mathcal{L}_{CE} + \lambda_1 \mathcal{L}_G + \lambda_2 \mathcal{L}_{Topo}
\end{equation}

where $\mathcal{L}_{CE} = -\frac{1}{T}\sum_{i=1}^T [y_i \log\hat{y}_i + (1-y_i)\log(1-\hat{y}_i)]$ is the cross-entropy loss for graph classification, and $\lambda_1,\lambda_2 \in (0,1)$ balance the loss components. This joint optimization enables simultaneous learning of local/global graph patterns through DDGformer, view-invariant features via contrastive learning, and topological signatures through persistent homology analysis.

\section{Experiments}

\subsection{Experimental Settings}
\textbf{Datasets}.Our model was evaluated on the Autism Brain Imaging Data Exchange (ABIDE)\cite{Martino} and Attention Deficit Hyperactivity Disorder (ADHD)\cite{ADHD}. ABIDE compiles data from 17 different acquisition sites, publicly sharing rs-fMRI and phenotypic data for 1112 subjects. In this work, the images analyzed were preprocessed using the Connectome Computation System (CCS). We utilized the ABIDE Preprocessed Connectomes Project (PCP) data, employing the configurable pipeline—Connectome Analysis (CPAC). The preprocessing steps included slice timing correction, motion correction, and voxel intensity normalization. After preprocessing, we obtained 871 high-quality MRI images with phenotypic information, including 403 ASD patients and 468 normal controls, with data from 17 different sites. The ADHD dataset employed in this study was sourced from New York University (NYU). The analyzed images were derived from the Preprocessed Connectomes Project (PCP) and underwent a comparable preprocessing procedure. Following preprocessing, high-quality MRI images were obtained from NYU, consisting of 96 normal controls and 81 ADHD patients. The Pearson Correlation Coefficients (PCCs) between each pair of brain regions were then calculated to serve as the initial features for these regions.

\textbf{Baseline Methods}.To comprehensively evaluate the performance of our proposed brain network diagnosis method (BrainCHEF), we compared it with existing state-of-the-art models. Specifically, these comparison methods can be divided into three major categories: traditional methods, including Functional Connectivity (FC) combined with Support Vector Machine (SVM)\cite{Soussia} and Random Forest (RF)\cite{Guo}; non-graph deep learning methods, including GroupINN\cite{Yan} and ASD-DiagNet\cite{Eslami}; and graph-structured deep learning methods, such as Spatial-Temporal Graph Convolutional Network (ST-GCN)\cite{Gadgil}, BrainGNN\cite{Li}, Multi-View Spectral Graph Convolutional Network (MVS-GCN)\cite{Wen1}, BrainGSL\cite{Wen2}, and FSL-BrainNet\cite{LiLanting}.

\textbf{Model Settings}.In this study, we adopted a five-fold cross-validation (5-fold Cross-Validation, CV) strategy to systematically evaluate the model performance. In each fold, 20\% of the samples were randomly selected from the training set as the validation set for hyperparameter tuning, ensuring the model's generalization capability across different data subsets. To reduce bias introduced by data partitioning, we performed five rounds of cross-validation, thereby enhancing the stability and reliability of the evaluation results. All experiments were implemented using the PyTorch and PyG (PyTorch Geometric) libraries. Specifically, the graph construction and aggregation were realized via PyG, while the graph transformer and the extraction of persistent homology features were implemented using PyTorch. The model was initialized using the Xavier initialization method, which effectively controls the output variance, thereby reducing numerical instability and ensuring the stability of network training. During the experiments, the Adam optimizer, which combines the benefits of RMSProp and Momentum, was employed for parameter updates. Moreover, a cosine annealing learning rate schedule was adopted to periodically adjust the learning rate, facilitating better convergence during model training.The parameters \( p_{e} \), \( p_{f} \), and \( p_{\tau} \) in Equations \(\ref{eq:edge_prob}\) and \(\ref{eq:masked_feat}\) were set based on the sparsity of the graph. For dense graphs, these parameters were set to 0.1, whereas for sparse graphs, they were set to 0.3. The damping factor \( d \) was set to the commonly used value of 0.85. The number of training epochs was set to 50, and the batch size was set to 32.

\textbf{Evaluation Metrics}.To comprehensively evaluate the performance of brain network classification models, we used four common evaluation metrics: classification accuracy (Accuracy, ACC), sensitivity (Sensitivity, SEN), specificity (Specificity, SPE), and the area under the receiver operating characteristic curve (Area Under the Curve, AUC). These metrics effectively reflect the model's ability to distinguish and predict accurately in brain network classification tasks.

\subsection{Performance Comparison}
\begin{table}[htbp]
  \centering
  \caption{Classification performance comparison between PHGCL-DDGformer and baseline models on ABIDE and ADHD-200 datasets}
  \label{tab:32}
  \setlength{\tabcolsep}{3pt} % Adjust column spacing for compactness
  \small % Set font size to small
  \begin{tabular}{lcccccccc}
    \toprule
    \textbf{Model} & \multicolumn{4}{c}{\textbf{ABIDE}} & \multicolumn{4}{c}{\textbf{ADHD-200}} \\
    \cmidrule(lr){2-5} \cmidrule(lr){6-9}
    & \textbf{ACC (\%)} & \textbf{AUC (\%)} & \textbf{SPEC (\%)} & \textbf{SEN (\%)} & \textbf{ACC (\%)} & \textbf{AUC (\%)} & \textbf{SPEC (\%)} & \textbf{SEN (\%)} \\
    \midrule
    SVM\textsuperscript{\cite{Soussia}} & 67.1$\pm$ 3.2 & 67.4$\pm$ 2.7 & 69.3$\pm$ 2.9 & 52.5$\pm$ 4.3 & 63.8$\pm$ 4.7 & 62.4$\pm$ 3.5 & 70.6$\pm$ 5.6 & 51.9$\pm$ 6.1 \\
    RF\textsuperscript{\cite{Guo}} & 64.5$\pm$ 3.7 & 62.2$\pm$ 2.8 & 62.9$\pm$ 4.1 & 46.4$\pm$ 5.8 & 61.6$\pm$ 4.8 & 56.1$\pm$ 4.3 & 59.8$\pm$ 4.8 & 56.3$\pm$ 8.0 \\
    GroupINN\textsuperscript{*}\cite{Yan} & 63.9 & 63.2 & 57.4 & 61.5 & 58.7 & 46.8 & 57.3 & 24.9 \\
    ASD-DiagNet\textsuperscript{*}\cite{Eslami} & 68.3 & 67.8 & 67.8 & 60.3 & 59.8 & 58.7 & 61.2 & 53.5 \\
    ST-GCN\textsuperscript{\cite{Gadgil}} & 57.3$\pm$4.0 & 51.7$\pm$2.9 & 48.9$\pm$5.6 & 54.8$\pm$5.4 & 52.5$\pm$7.6 & 53.9$\pm$8.4 & 55.6$\pm$8.2 & 49.1$\pm$12.9 \\
    BrainGNN\textsuperscript{\cite{Li}} & 61.8$\pm$ 2.2 & 60.8$\pm$ 2.5 & 60.8$\pm$ 2.6 & 61.7$\pm$ 3.7 & 61.0$\pm$ 3.2 & 57.8$\pm$ 2.2 & 64.1$\pm$ 4.2 & 54.7$\pm$ 6.1 \\
    FSL-BrainNet\textsuperscript{*}\textsuperscript{\cite{LiLanting}} & 69.8 & \textbf{75.9} & 68.8 & 70.9 & 61.6 & \textbf{63.7} & 61.8 & 53.1 \\
    MVS-GCN\textsuperscript{*}\textsuperscript{\cite{Wen1}} & 69.8 & 69.1 & 63.1 & 70.2 & 62.2 & 61.1 & 66.4 & 56.7 \\
    BrainGSL\textsuperscript{*}\textsuperscript{\cite{Wen2}} & 70.4 & 71.1 & 69.5 & 70.6 & 62.3 & 61.5 & 66.3 & 56.2 \\
    PHGCL-DDGformer & \textbf{70.9$\pm$4.1} & 73.5$\pm$5.0 & \textbf{70.9$\pm$2.4} & \textbf{74.8$\pm$8.0} & \textbf{67.8$\pm$4.3} & 63.3$\pm$1.8 & \textbf{68.8$\pm$3.5} & \textbf{61.1$\pm$5.8} \\
    \bottomrule
    \multicolumn{9}{l}{\footnotesize \textit{Note}: * indicates results reproduced from original publications.}
  \end{tabular}
\end{table}
 All models were evaluated using five-fold cross-validation on both ABIDE and ADHD-200 datasets, with detailed results presented in Table~\ref{tab:32}.The PHGCL-DDGformer demonstrates superior performance by effectively extracting graph-structured information while employing Gaussian decay functions to constrain the attention mechanism in graph transformers, thereby addressing the over-globalization problem in homogeneous graphs. This architecture enables more precise feature extraction and information integration from complex brain networks.Compared with traditional machine learning methods (SVM and RF), PHGCL-DDGformer achieves significant improvements in both accuracy and sensitivity, reaching 70.9\% and 74.8\% on ABIDE, and 67.8\% and 61.1\% on ADHD-200 respectively. These results demonstrate our model's capability to overcome the limitations of conventional approaches in modeling complex brain networks and extracting latent discriminative features.When compared with non-graph deep learning methods (GroupINN and ASD-DiagNet), PHGCL-DDGformer shows remarkable advantages in both specificity and accuracy, confirming that graph-structured modeling better captures the intrinsic properties of brain networks by treating them as holistic graphs.Among graph-based deep learning approaches, PHGCL-DDGformer achieves state-of-the-art performance on most metrics. On ABIDE, it outperforms all competitors in accuracy while attaining comparable AUC (second only to FSL-BrainNet). Notably, on ADHD-200, our model achieves superior AUC compared to FSL-BrainNet. These results validate that PHGCL-DDGformer provides more comprehensive understanding of brain network complexity and better identifies disease-relevant biomarkers for classification tasks.

\subsection{Parameter Analysis}
\subsubsection{Impact of Sparsity Ratio on Model Performance}

\begin{figure}[htbp]
    \centering
    \subfigure[ABIDE dataset]{
        \label{Node.sub.1}
        \includegraphics[width=0.46\linewidth]{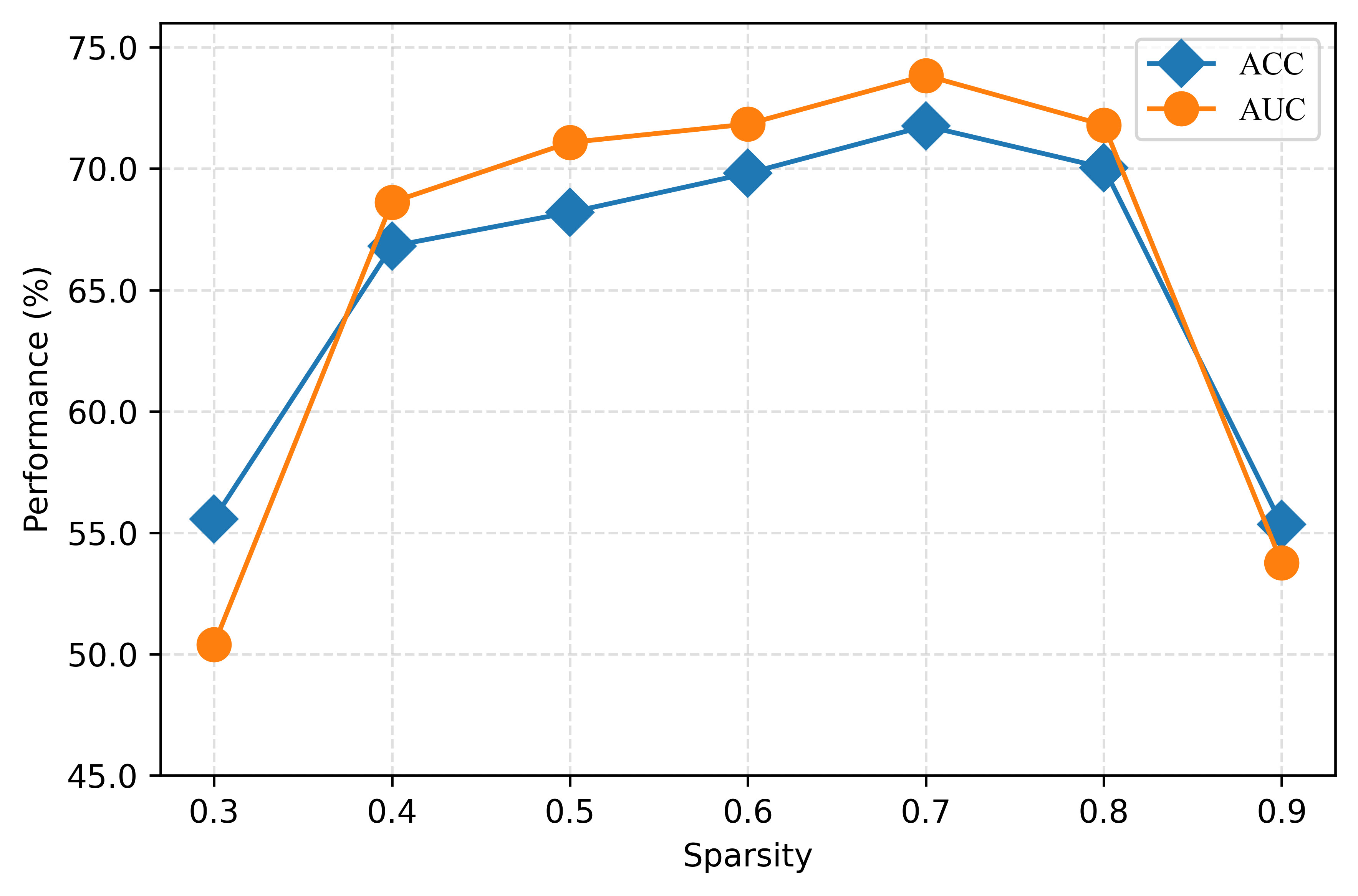}}
    \quad % Default spacing between subfigures is small; this command increases width
    \subfigure[ADHD-200 dataset]{
        \label{Node.sub.2}
        \includegraphics[width=0.46\linewidth]{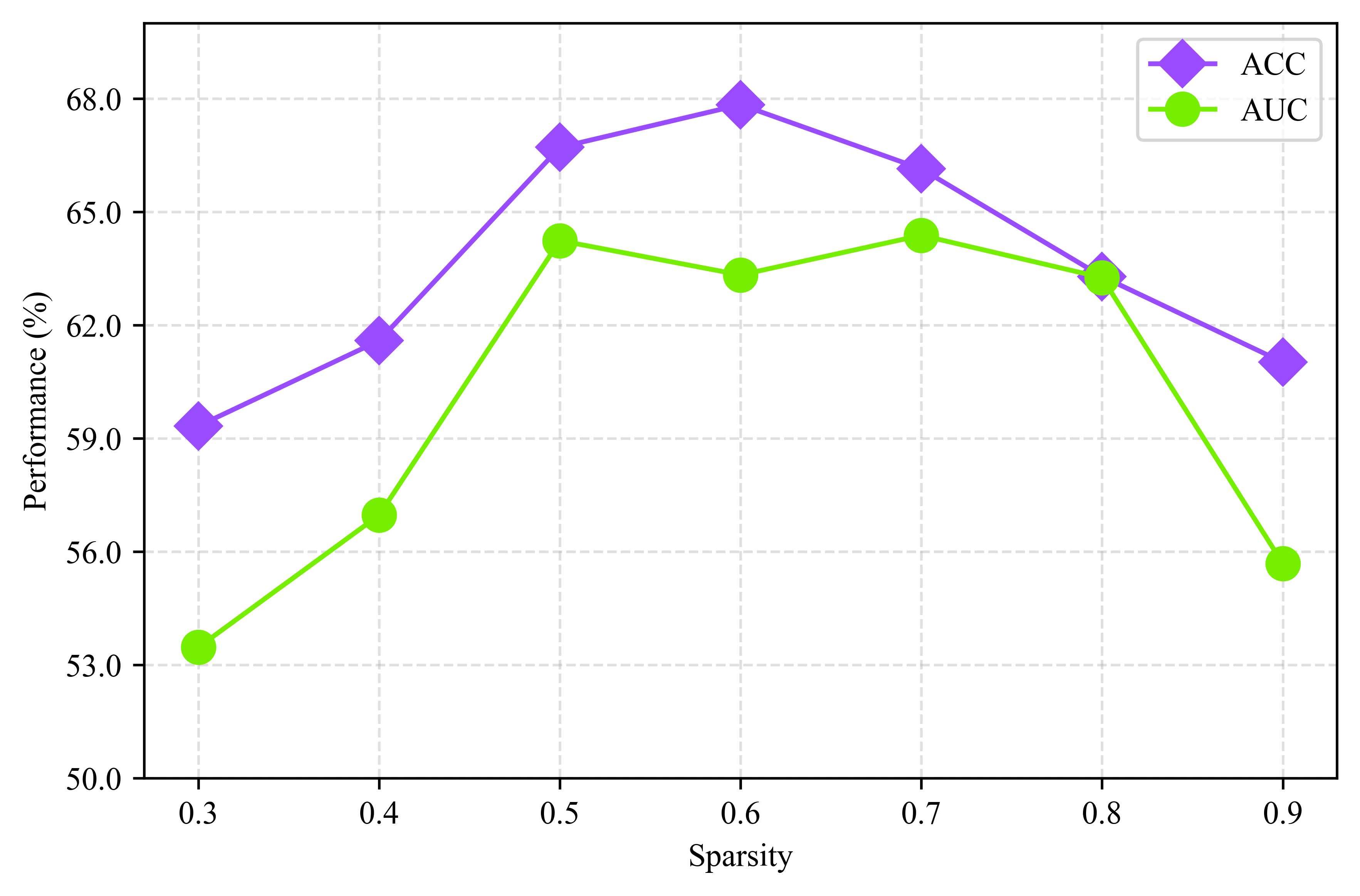}}
    \caption{Impact of correlation matrix sparsity ratio on PHGCL-DDGformer model performance}
    \label{fig:thresholds}
\end{figure}

In graph structure binarization, different thresholds can control the sparsity ratio of brain network matrices, thereby affecting model expressiveness. This study systematically investigates the influence mechanism of threshold settings on the performance of the PHGCL-DDGformer model. Specifically, we conducted comprehensive parameter tuning experiments by adjusting the threshold range to ensure sparsity ratios within \{0.1, 0.2, ..., 0.7\}. The experiments used Area Under the Receiver Operating Characteristic Curve (AUC) and Accuracy (ACC) as evaluation metrics, with results shown in Figure \ref{fig:thresholds}. Model performance demonstrated a significant trend of first increasing and then decreasing, with maximum AUC reaching 73.5\% and maximum ACC reaching 70.9\%. 

Experimental results indicate that as sparsity ratio increases, model performance shows an initial improvement followed by decline. Specifically, for the ABIDE dataset at lower sparsity (0.3), the network retains more connections resulting in denser brain networks that may contain more noise or redundant information. At higher sparsity (0.9), the loss of key node correlation information limits the model's ability to capture high-order structural features. With moderate sparsity, the binarization process preferentially preserves edges with stronger correlations in the brain network, effectively reducing the impact of noisy or redundant connections while retaining the most reliable and information-rich connections. Therefore, the ABIDE dataset selected a sparsity ratio of 0.3 as the optimal configuration, while the ADHD-200 dataset chose 0.4.

\subsubsection{Impact of Encoding Layers \(L\)}
The number of encoding layers \(L\) reflects the model's capability in feature extraction and information fusion, serving as a crucial parameter in the PHGCL-DDGformer model. Analyzing the influence of different \(L\) values on model performance can effectively optimize the model and investigate its intrinsic mechanisms. Experimental results are shown in Table \ref{tab:EN}.The variation in \(L\) significantly affects model performance. When \(L = 1\), the single encoding layer demonstrates suboptimal performance due to insufficient extraction of global and local features from complex graph structures. When \(L = 2\), the dual-layer encoding achieves substantially improved performance by effectively integrating global and local information while maintaining relatively low complexity. However, when \(L = 3\), the model performance declines as increased complexity leads to overfitting on training data, consequently reducing generalization capability.
\begin{table}[htbp]
  \centering
  \caption{Impact of encoding layers $L$ on PHGCL-DDGformer model performance}
  \label{tab:EN}
  \small
  \setlength{\tabcolsep}{6pt}
  \renewcommand{\arraystretch}{1.2}
  \begin{tabular}{lcccc}
    \toprule
    \textbf{Layers ($L$)} & \textbf{ACC (\%)} & \textbf{AUC (\%)} & \textbf{SPEC (\%)} & \textbf{SEN (\%)} \\
    \midrule
    $L=1$      & 64.8 $\pm$ 4.2 & 65.4 $\pm$ 4.6 & 64.1 $\pm$ 7.8 & 70.4 $\pm$ 11.0 \\
    $L=2$      & 70.9 $\pm$ 4.1 & 73.5 $\pm$ 5.0 & 70.9 $\pm$ 2.4 & 74.8 $\pm$ 8.0 \\
    $L=3$      & 68.4 $\pm$ 4.4 & 72.6 $\pm$ 1.6 & 69.4 $\pm$ 6.1 & 72.2 $\pm$ 13.2 \\
    \bottomrule
    \end{tabular}
\end{table}

\subsubsection{Impact of Contrastive Constraint Hyperparameters $\lambda_{1}$ and $\lambda_{2}$}

\begin{figure}[htbp]
    \centering
    \subfigure[ACC (left) and AUC (right) values on ABIDE dataset]{
        \includegraphics[width=0.4\textwidth]{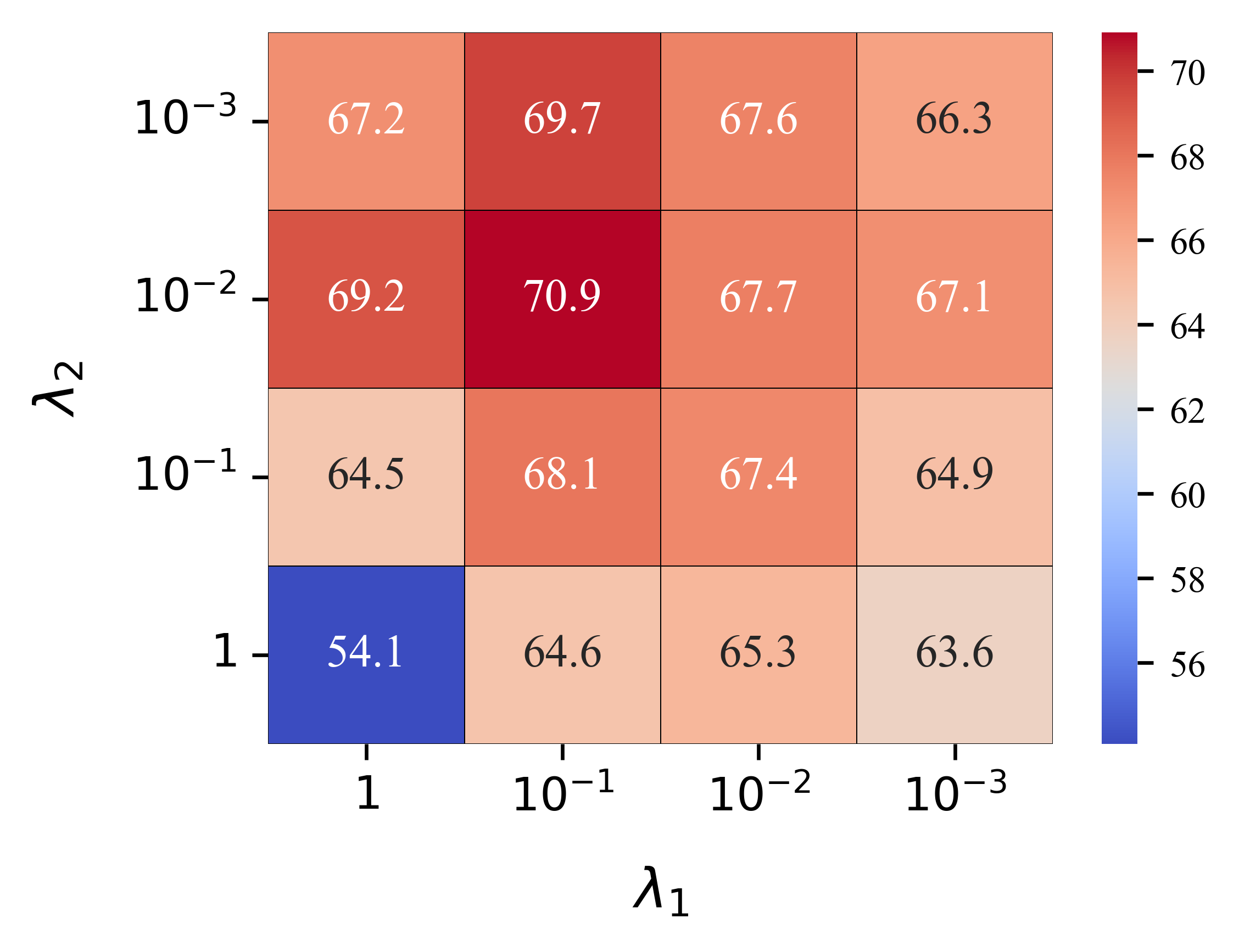}
        \includegraphics[width=0.4\textwidth]{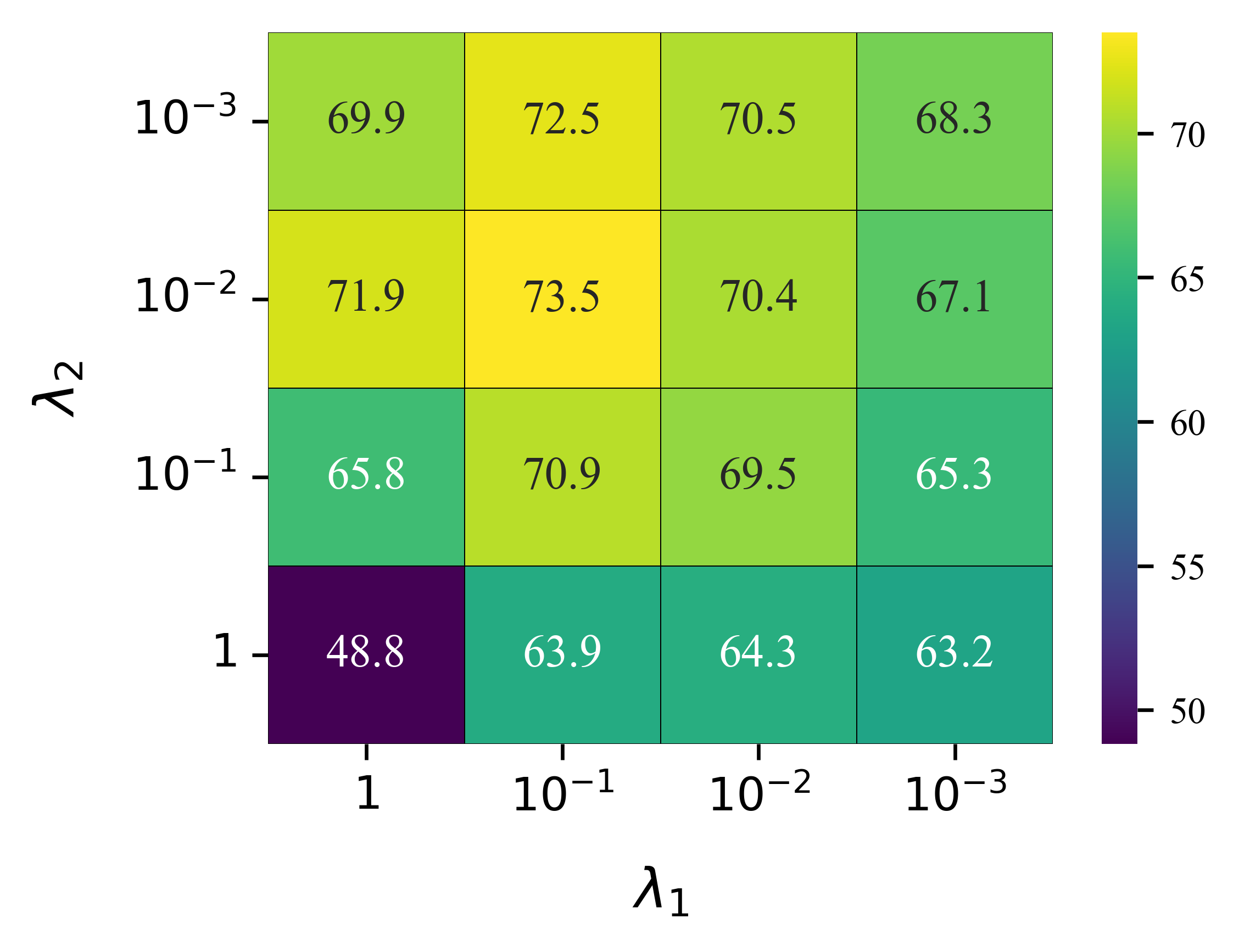}
        \label{fig:19a11}
    }
    \subfigure[ACC (left) and AUC (right) values on ADHD-200 dataset]{
        \includegraphics[width=0.4\textwidth]{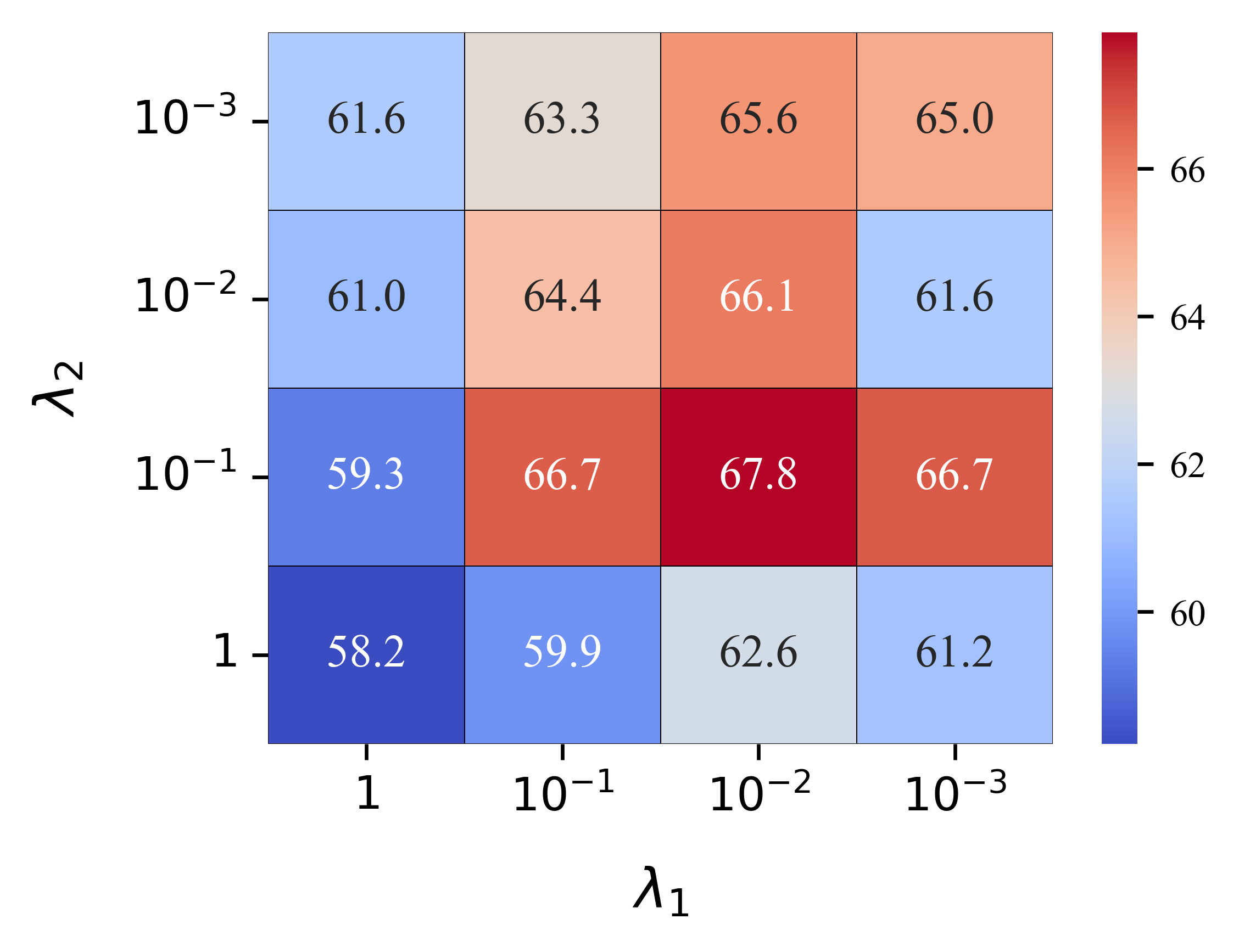}
        \includegraphics[width=0.4\textwidth]{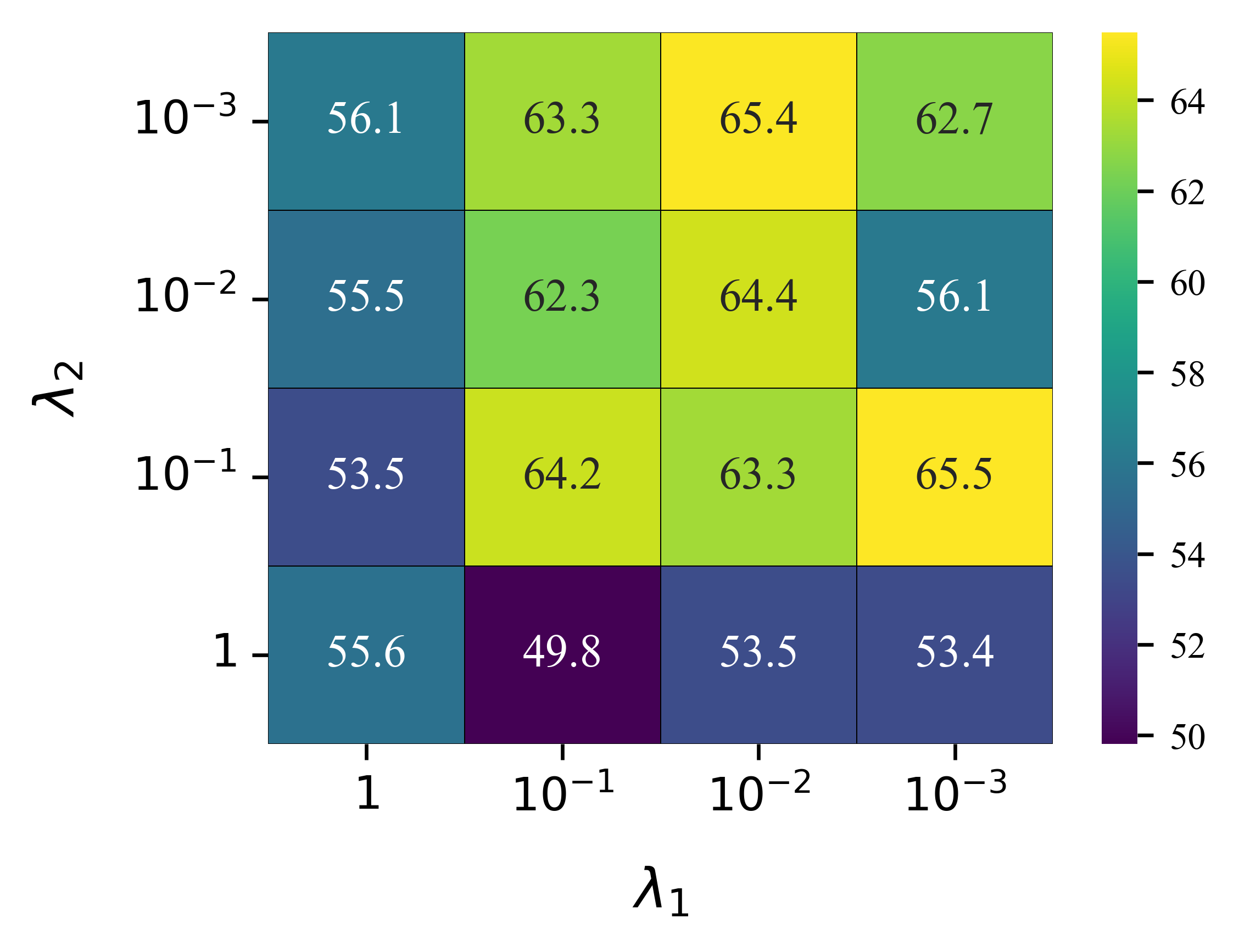}
        \label{fig:19b11}
    }
    \caption{Impact of contrastive constraint hyperparameters $\lambda_{1}$ and $\lambda_{2}$ on PHGCL-DDGformer model performance}
    \label{fig:Hyperparameter}
\end{figure}

The PHGCL-DDGformer model (see Equation \ref{eqqqq}) introduces hyperparameters $\lambda_{1}$ and $\lambda_{2}$ to precisely regulate graph-level contrastive constraints. To comprehensively analyze their impact on model performance, we systematically adjusted $\lambda_{1}$ and $\lambda_{2}$ values across the range \(\{10^{-3}, 10^{-2}, 10^{-1}, 1\}\) using training and validation sets, capturing behavioral variations under different constraint intensities. Experimental results are collectively presented in Figure \ref{fig:Hyperparameter}, providing visual support for subsequent analysis.Results demonstrate significant performance degradation when $\lambda_{1}$ and $\lambda_{2}$ adopt larger values (e.g., $\lambda_{1}=1$ and $\lambda_{2}=1$), indicating excessive contrastive constraints prevent model convergence to an optimal solution space, substantially impairing learning capability. Conversely, weak contrastive constraints (e.g., $\lambda_{1}=10^{-3}$ and $\lambda_{2}=10^{-3}$) also yield suboptimal performance, typically inferior to moderately strong constraints (e.g., $\lambda_{1}=10^{-2}$ and $\lambda_{2}=10^{-2}$). This suggests appropriate contrastive constraints are crucial for learning discriminative features. 

In summary, insufficient attention to brain contrast views prevents full exploitation of deep structural information, consequently hindering satisfactory classification performance. Optimal configurations were achieved with \(\lambda_1 = 10^{-1}\) and \(\lambda_2 = 10^{-2}\) for ABIDE dataset (yielding peak ACC), while \(\lambda_1 = 10^{-2}\) and \(\lambda_2 = 10^{-1}\) proved optimal for ADHD-200 dataset.

\subsection{Ablation Study}
\begin{table}[htbp]
  \centering
  \caption{Ablation study results of PHGCL-DDGformer model}
  \setlength{\tabcolsep}{1mm}
  \begin{tabular}{ccccccc}
    \toprule
    Model & Ada & DDformer & GCL & Topo & ACC(\%) & AUC(\%) \\
    \midrule
    GCN   & & & &    & 66.8 & 68.6 \\
    DDformer & &\checkmark & &    & 67.3 & 67.9 \\
    GCN + Ada    & \checkmark & & &    & 64.7 & 66.2 \\
    DDformer+Ada  & \checkmark & \checkmark& &    & 65.5 & 65.6 \\
    PHGCL w/o Topo  & \checkmark & & \checkmark&   & 67.7 & 69.4 \\
    PHGCL-DDGformer w/o Topo  & \checkmark & \checkmark&\checkmark &    & 69.8 & 71.6 \\
    PHGCL  & \checkmark & &\checkmark & \checkmark   & 68.8 & 70.1 \\
    PHGCL-DDGformer  & \checkmark & \checkmark& \checkmark&  \checkmark  & 70.9 & 73.5 \\
    \bottomrule
    \label{tab:PHGCL}
  \end{tabular}
\end{table}

In the PHGCL-DDGformer architecture, the graph augmentation module preserves crucial structural and attribute information, while the DDformer component effectively extracts both local and global features along with complex interaction patterns from brain networks. The graph contrastive learning module further integrates local node/edge information with higher-order structural patterns including subgraph configurations, network motifs, and global topological invariants, thereby enhancing brain network representation. To investigate each component's contribution, we conducted comprehensive ablation studies on the ABIDE dataset, with results presented in Table \ref{tab:PHGCL}.The results demonstrate that directly incorporating graph augmentation degrades performance, potentially due to disruption of critical structural information during the augmentation process. The DDformer module significantly improves model capability by enabling nodes to aggregate information from both distant and local neighbors, enhancing comprehension of complex brain network structures. The graph contrastive learning strategy provides substantial performance gains by leveraging complementary information across different views, effectively expanding the sample space to improve parameter estimation accuracy. Finally, the incorporation of higher-order topological information proves essential for optimal model performance.

\subsection{Interpretability Analysis}
\begin{figure}[htbp]
    \centering
    \includegraphics[scale=0.06]{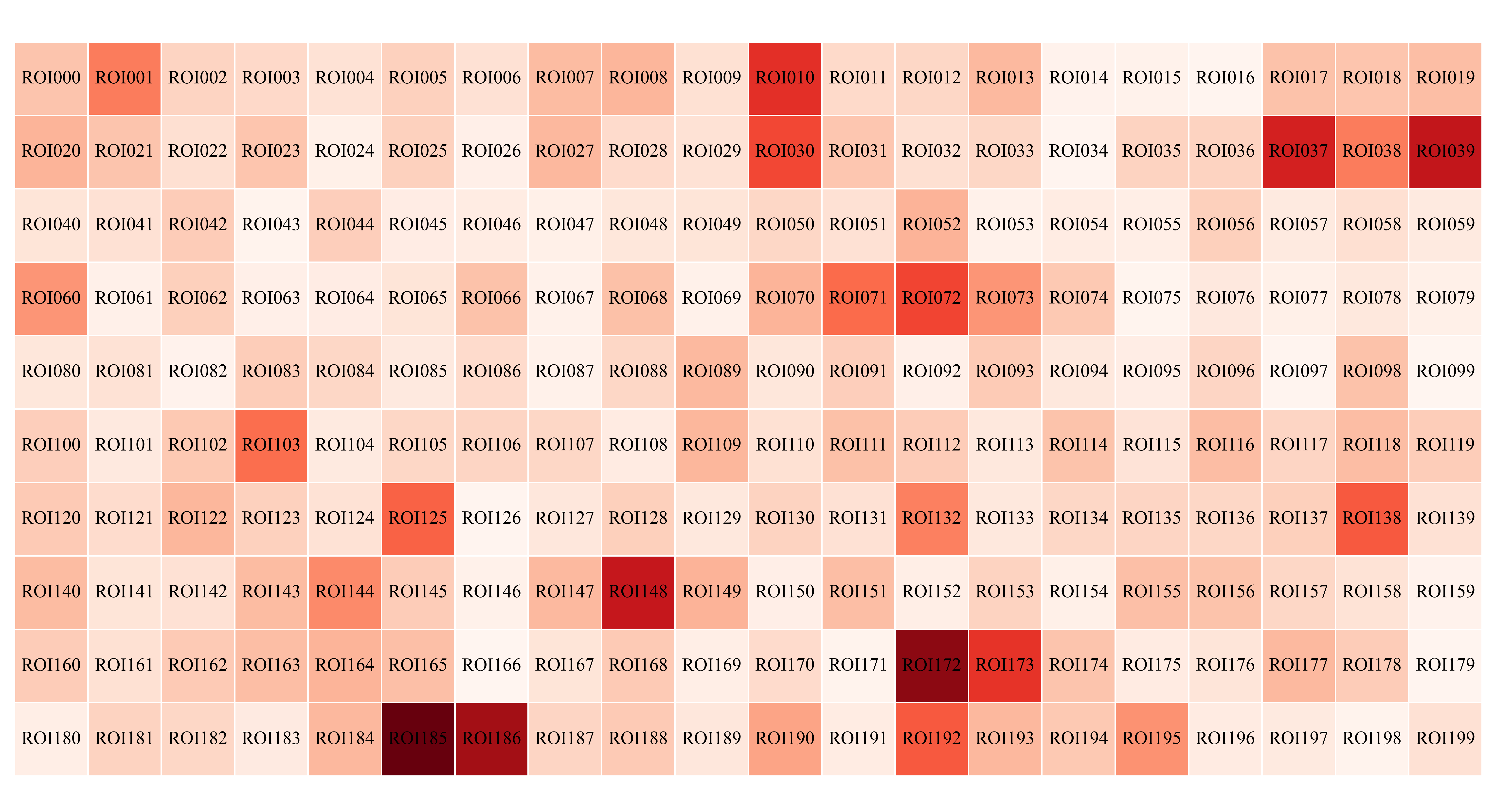}
    \caption{Visualization of brain region attention coefficients under CC200 template: Color intensity reveals autism spectrum disorder (ASD) relevance}
    \label{fig:ROI}
\end{figure}

To investigate PHGCL-DDGformer's classification mechanism for brain networks, we employed attention graph pooling to score regions of interest (ROIs) on the ABIDE dataset, identifying disease-relevant brain regions as potential biomarkers (visualized in Figure \ref{fig:ROI}). Since Craddock200 parcellation defines ROIs through clustering, we conducted cross-referencing analysis with the widely-used AAL atlas. Results identified the top three ASD-associated ROIs: left middle temporal gyrus (Temporal\_Mid\_L), left angular gyrus (Angular\_L), and left middle occipital gyrus (Occipital\_Mid\_L).These findings align with prior ASD research: the angular gyrus associates with reading comprehension and attention \cite{seghier2013angular}, and exhibits social interaction deficits \cite{lord2022lancet}; the middle temporal gyrus, crucial for social network processing, shows altered functional connectivity in ASD patients \cite{bachevalier1994medial}; the middle occipital gyrus mediates object/face recognition, where ASD patients demonstrate functional connectivity differences; the superior frontal gyrus, as a core default mode network component, relates to social behavior with impaired connectivity in ASD \cite{xu2020specific}; the middle frontal gyrus links to ASD's neurocognitive pathophysiology, where abnormal cortical volume/connectivity may cause working memory deficits. These results validate PHGCL-DDGformer's effectiveness in identifying disease-relevant brain regions.

\section{Conclusion}
This section presented PHGCL-DDGformer, a novel graph contrastive learning framework that significantly improves brain network classification accuracy and generalizability. Addressing data scarcity and weak supervision in brain networks, we developed: (1) attribute masking and edge perturbation modules for graph augmentation; (2) DDGformer for local/global feature extraction and complex interaction modeling; and (3) graph contrastive learning modules.

Despite its superior performance, limitations remain: the augmentation module may not perfectly preserve critical features; higher-order interactions for capturing co-activation patterns require further optimization; and computational complexity needs reduction. Future work will focus on: (1) advanced brain network modeling via higher-order structures (e.g., hypergraphs/simplicial complexes); (2) computational efficiency improvements; and (3) broader validation across neurological disorders.
%Bibliography
\bibliographystyle{unsrt}  
\bibliography{references}

\end{document}